\documentclass[11pt,a4paper]{article}
\usepackage[hyperref]{acl2019}
\usepackage{times}
\usepackage{latexsym}

\usepackage{url}
\usepackage{graphicx}
\graphicspath{{50Images/}}
\usepackage{array,multirow}
\usepackage{url}
\usepackage{soul}
\usepackage{amsmath}
\usepackage{color}
\usepackage[normalem]{ulem}
\usepackage{makecell}

\aclfinalcopy 
\def\aclpaperid{78} 

\title{Detecting Machine-Translated Text using Back Translation}

\author{Hoang-Quoc Nguyen-Son$^\textsuperscript{1}$, Tran Phuong Thao$^\textsuperscript{2}$, Seira Hidano$^\textsuperscript{1}$, and Shinsaku Kiyomoto$^\textsuperscript{1}$ \\
  $^\textsuperscript{1}$KDDI Research, Inc. \\
  2-1-15, Ohara, Fujimino, Saitama, 356-8502, Japan\\
  {\tt \{ho-nguyen, se-hidano, kiyomoto\}@kddi-research.jp} 
  \\
  $^\textsuperscript{2}$The University of Tokyo \\
  7-3-1, Hongo, Bunkyo, Tokyo, 113-8656, Japan\\
  {\tt tpthao@yamagula.ic.i.u-tokyo.ac.jp} 
  }
  
\date{}

\begin{document}
\maketitle
\begin{abstract}
Machine-translated text plays a crucial role in the communication of people using different languages.
However, adversaries can use such text for malicious purposes such as plagiarism and fake review.
The existing methods detected a machine-translated text only using the text's intrinsic content, but they are unsuitable for classifying the machine-translated and human-written texts with the same meanings.
We have proposed a method to extract features used to distinguish machine/human text based on the similarity between the intrinsic text and its back-translation.
The evaluation of detecting translated sentences with French shows that our method achieves 75.0\% of both accuracy and $F$-score. 
It outperforms the existing methods whose the best accuracy is 62.8\% and the $F$-score is 62.7\%.
The proposed method even detects more efficiently the back-translated text with 83.4\% of accuracy, which is higher than 66.7\% of the best previous accuracy.
We also achieve similar results not only with $F$-score but also with similar experiments related to Japanese.
Moreover, we prove that our detector can recognize both machine-translated and machine-back-translated texts without the language information which is used to generate these machine texts.
It demonstrates the persistence of our method in various applications in both low- and rich-resource languages.
\end{abstract}

\section{Introduction}

Nowadays, cross-language communication among people plays an important role in modern life.
It opens great opportunities in various fields such as entertainment, e-commerce, career, etc.
In this communication, a machine translator is an essential component.
Moreover, the translator can also support other mutual interactions among machines and between a human with a machine.
For example, a new AI system can be built from the other mature systems, which are operated in another language.
In another example, the cutting-edge smart devices such as Apple Siri and Google Home have already supported multiple languages via translators.

However, the main problem of using translation is that it can lead to misunderstanding due to the diversity of language usages such as slang, idiom, dialect, etc.
In another problem, adversaries can take advantage of translators to generate paraphrasing texts for malicious purposes, for example, plagiarism~\cite{jones2015back} and style transfer~\cite{prabhumoye2018style}.
Spreading such artificial texts can seriously reduce the reputation of the original texts which are created from human society.
Therefore, it is crucial to develop a detector for determining whether a text is written by a human or generated by a translator.

Many researchers have interested in detecting machine-translated text.
The most common methods are based on the $N$-gram model~\cite{aharoni2014automatic,arase2013machine,nguyen2017detecting} to measure the fluency of text.
On the other hand, the structure of the parsing tree is exploited to recognize the machine-generated texts~\cite{li2015machine}.
Moreover, the different word usages in human and machine texts lead to the differences in their word distributions~\cite{nguyen2017identifying}.
Other researchers prove that the coherence of the human-written text is better than the machine-translated one~\cite{nguyen2018identifying,nguyen2019detecting}.
Beyond detecting machine-translated text, artificial fake reviews and papers are also recognized by readability~\cite{juuti2018stay} and duplicate patterns~\cite{labbe2013duplicate}, respectively.
The limitation in all existing methods above is that they only analyze the intrinsic contents of machine-generated texts but ignore the original processes which are used to produce the texts.

\begin{figure}[t]
\centering
\includegraphics[]{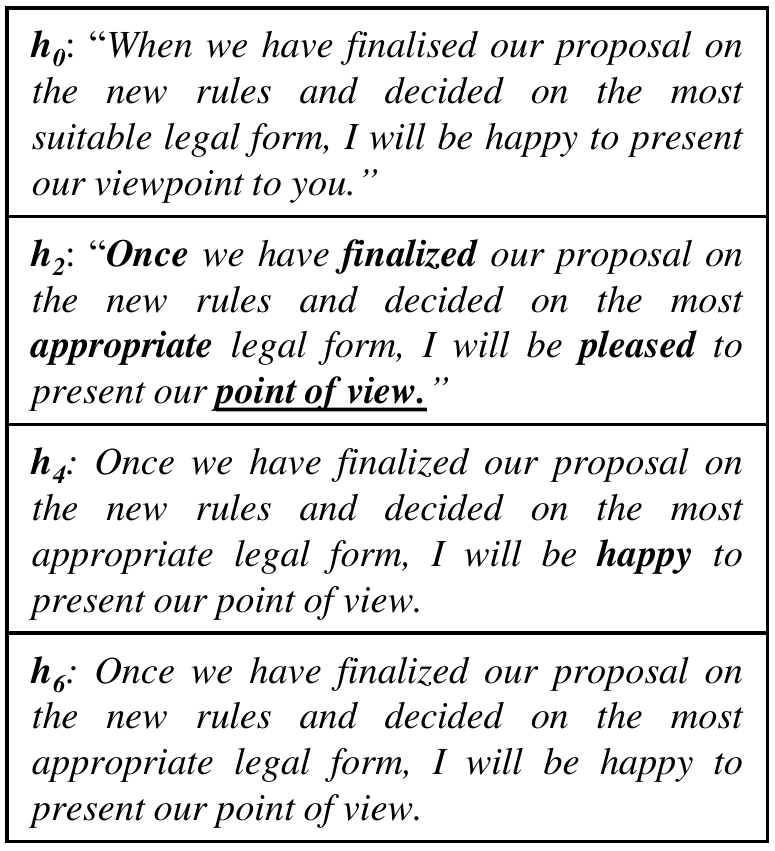}
\caption{The variants of repeatedly using back-translations.}
\label{Fig_0_Hypothesis}
\end{figure}

Our idea based on the fact that the processing on original data often produces more variations than that on modified data.
For example, in the field of image, equalizing histogram on an original image makes the much larger change than that on a balanced image, which has already equalized before.
In the field of text, we also have a similar phenomenon.
More particularly, we conduct a random example on an original sentence $h_0$ from European parallel corpus\footnote{\url{https://www.statmt.org/europarl/}} as shown in Figure~\ref{Fig_0_Hypothesis}.
$h_0$ is translated to French and then re-translated to English to create the back-translation denoted as $h_2$ where the subscript $2$ represents to the number times of transitions applied from $h_0$.
The other back-translations $h_4$ and $h_6$ are generated in the same manner of $h_2$.
The variation between a back-translation with its origin is highlighted in bold with word usage and in underline with structure.
The back-translation reach the saturation in $h_6$ with no change.
Among back-translations, $h_2$ has the largest variations with seven positions in the word usage and three positions in the structure.
The variations are remarkably reduced in the next back-translation $h_4$ with only one position in the word usage and nothing in $h_6$.
The example demonstrates that the earlier generations have a higher number of variations than the latter ones.

\begin{figure*}[t]
\centering
\includegraphics[]{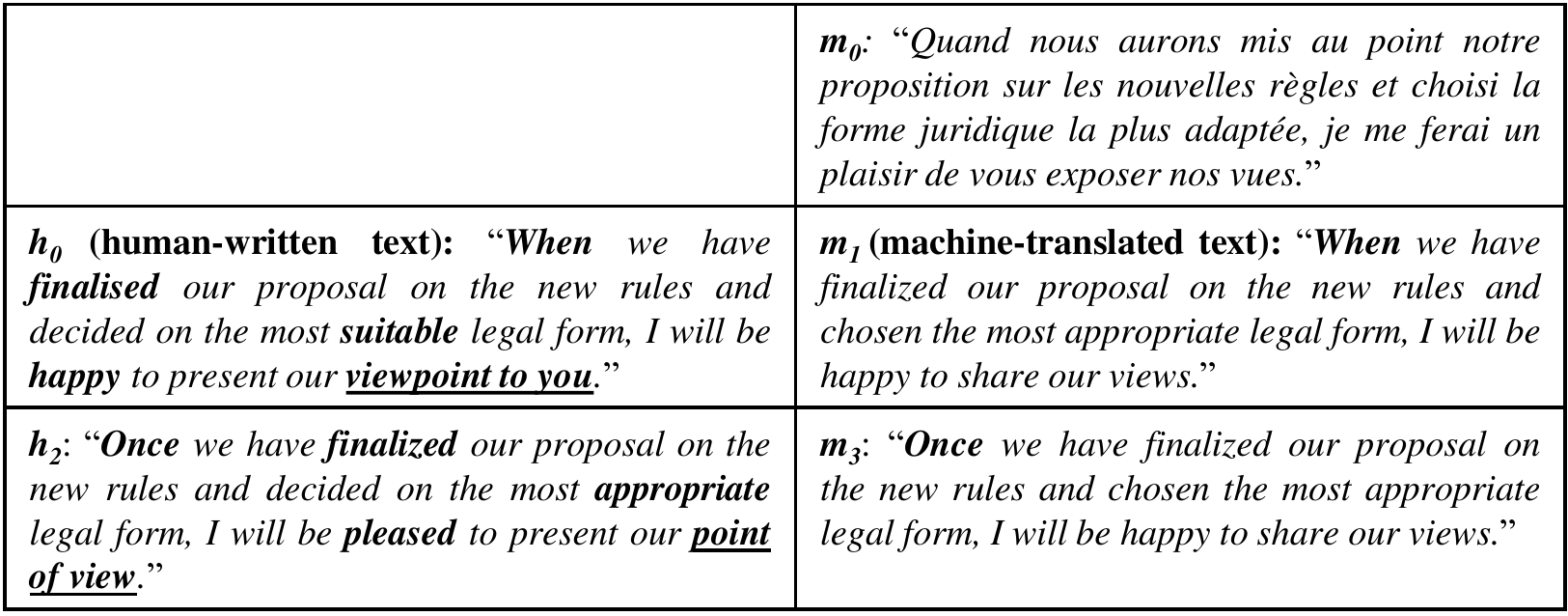}
\caption{Human-written vs machine-translated text.}
\label{Fig_1_Example}
\end{figure*}

We check our findings on machine-translated text detection.
More particularly, we picked up the English-French pair $\{h_0,m_0\}$ in the parallel corpus in which $h_0$ is analyzed above.
While $h_0$ is considered as the human-written sentence, $m_0$ is translated to English by Google for generating a machine sentence $m_1$ as shown in Figure~\ref{Fig_1_Example}.
We then generate the two back-translation versions $h_2$ and $m_3$ using French as the intermediate language.
While $h_2$ is translated in two times from the origin $h_0$, $m_3$ is generated after three from $m_1$.
At the result, $h_2$ has more variations with $h_0$ in word usage than $m_3$ with $m_1$.
Moreover, the structure in $h_2$ is slightly changed whereas in $m_3$ is preserved.
It demonstrates that the differences in back-translation can be used to distinguish human-written with machine-translated text.

In this paper, we have proposed a method using back-translation to detect machine-translated text. 
Our contributions are listed as below:

\begin{itemize}
    \item 
    We explore the variant of a text when repeatedly back-translated in the same translator.
    In particular, the text is invariant after certain times of back-translating.
    Moreover, the earlier back-translations produce the larger variants than the later ones.
    \item 
    We measure the variant by calculating the similarity between the text and its back-translation using BLEU scores.
    \item 
    We suggest using a classifier with these scores to determine whether the text is translated by a machine or written by a human.
\end{itemize}

We randomly selected 2000 English-French sentence pairs from the European corpus for evaluation.
While the English was considered as the human-written text, the French was translated to English using Google and is represented for the machine-translated text.
Our method achieves both accuracy and $F$-score as 75.0\%.
It outperforms previous methods with the best accuracy as 62.8\% and $F$-score as 62.7\%.
The similar experiment was conducted with back-translation detection.
More specifically, we randomly chose 2000 sentiment sentences including 1000 positives and 1000 negatives from a Stanford Treebank corpus\footnote{\url{http://nlp.stanford.edu/~socherr/stanfordSentimentTreebank.zip}}.
We then generated the machine back-translated text using French as the intermediate language.
Our performance gives 83.4\% of both accuracy and $F$-score that is better than the best previous work's accuracy and $F$-score as 66.7\% and 63.7\%, respectively.
We conducted further experiments with Japanese and reach similar results.
It demonstrates the persistence of the proposed method in various tasks in both low- and rich-resource languages.

The rest of the paper is organized as follow.
Section~\ref{section:related_work} describes some main previous methods of detecting machine-translated and other machine-generated texts.
The proposed method is presented in Section~\ref{section:proposed_method}.
The experimental results are shown in Section~\ref{section:evaluation}.
Finally, we summarize some main key points and mention future work in Section~\ref{section:conclusion}.

\section{Related Work}
\label{section:related_work}

\subsection{Machine Translation Detection}

The previous methods for detecting machine-translated text can be split into four groups.

\paragraph{$N$-gram model}
This model is commonly used to estimate the fluency of continuous words.
Researchers have suggested additional features to support the original model.
For example, \newcite{arase2013machine} estimated the fluency of non-continuous words by sequential pattern mining.
They can extract fluent human patterns (e.g., ``\textit{not only * but also},'' and ``\textit{more * than}'') comparing with weird machine patterns (e.g., ``\textit{after * after the},'' ``\textit{and also * and}'').
On the other hand, \newcite{aharoni2014automatic} combined the POS $N$-gram model with functional words, which abundantly occur in the machine-translated text.
\newcite{nguyen2017detecting} also integrated the word $N$-gram model with noise features for detecting translation in online social networking (OSN) messages.
Such specific features often occur in human messages such as misspelling and spoken words or in machine messages, for example, untranslated words.
However, these noises frequently appear in the OSN messages more than others.

\paragraph{Parsing tree}
\newcite{li2015machine} used the syntactic parsing tree for classifying human and machine sentences.
They claim that the structure of a human parsing is more balancing than that of a machine.
They thus extracted balancing-based features such as the ratio between left and right nodes in both general and main continents.
The limitation of this approach is that it ignores the semantic meaning of the text.

\paragraph{Word distribution}
The usage of words in the human text often complies the Zipfian law, which indicates the topmost frequent words double the second, three times the third, etc.
\newcite{nguyen2017identifying} use this law for detecting machine translated document.
Furthermore, they extracted useful humanity text including idiom, clich\'{e}, ancient, and dialect phrases.
They also estimated the relationships among certain phrases based on co-reference resolutions.
These features only work well on a large text in which the word distribution is more stable and additional features appear more.

\paragraph{Coherence} 
Although the machine-translated text can preserve the meaning, the coherence of such text is still low.
Some researchers have measured the coherence to distinguish the machine text with the human text.
For example, \newcite{nguyen2018identifying} matched similar words between two sentences in a paragraph.
The similarity between two matched words is used to estimate the coherence.
In another work, \newcite{nguyen2019detecting} broadened the matching on any words in the paragraph in both within and across sentences.
However, the coherence is tight in a paragraph but is downgraded in other levels such as sentence and document.

\subsection{Other Machine-Generated Text Detection}

Many other machine-generated texts support for malicious purposes such as paper generation and fake review.
\newcite{labbe2013duplicate} prove that artificial papers are produced by using abundant duplicated words and phrases.
Therefore, they suggested an inter-textual distance to estimate the similarity between two word distributions and used the distance to recognize the machine-generated text.
In fake review detection, \newcite{juuti2018stay} extracted features from thirteen readability metrics.
Moreover, they used $N$-gram models for various text components including words, simple POS, detailed POS and syntactic dependency.
The duplicated usages of word distribution and $N$-gram model indicate high relevant between machine-translated and other machine-generated texts detection.

\section{Proposed Method}
\label{section:proposed_method}

The schema of the proposed method includes three steps as shown in Figure~\ref{Fig_2_Proposed_schema}:

\begin{figure}[t]
\centering
\includegraphics[]{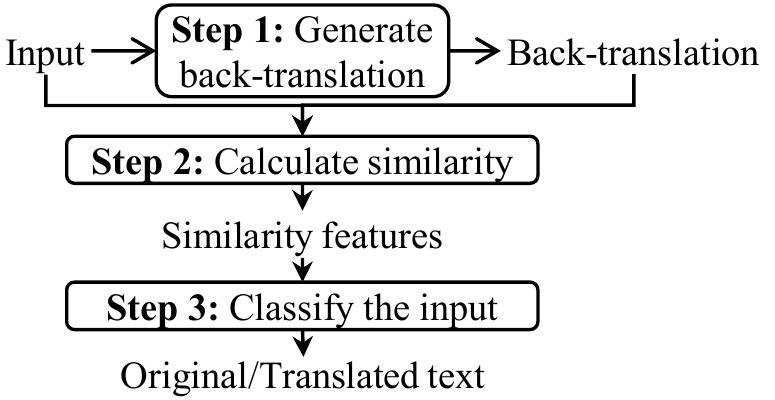}
\caption{The proposed schema for detecting machine-translated text.}
\label{Fig_2_Proposed_schema}
\end{figure}

\begin{itemize}
    \item 
    
    \textbf{Step 1 (\textit{Generate back-translation})}: The Google Translation is used to generate the back-translation of the input text.
    
    \item
    
    \textbf{Step 2 (\textit{Calculate similarity})}: The similarity between the input text and its back-translation is measured on the basis of BLEU scores.
    
    \item
    
    \textbf{Step 3 (\textit{Classify the input})}: The similarity features are used to determine whether the input text is written by a human or generated by a machine.
\end{itemize}

The following subsections describe the step-by-step of the proposed method.

\subsection{Generating Back-Translation (Step 1)}

The input text in the original language is translated into an intermediate language, which is different from the original one.
The translated version is then re-translated back to the original language.
The final translation is called as back-translation.
In this paper, we use Google as a translator.
In Figure~\ref{Fig_1_Example}, the back-translations $h_2$ and $m_3$ are generated from the human text $h_0$ and machine text $m_1$ respectively with the intermediate language, French.

\begin{figure*}[t]
\centering
\includegraphics[]{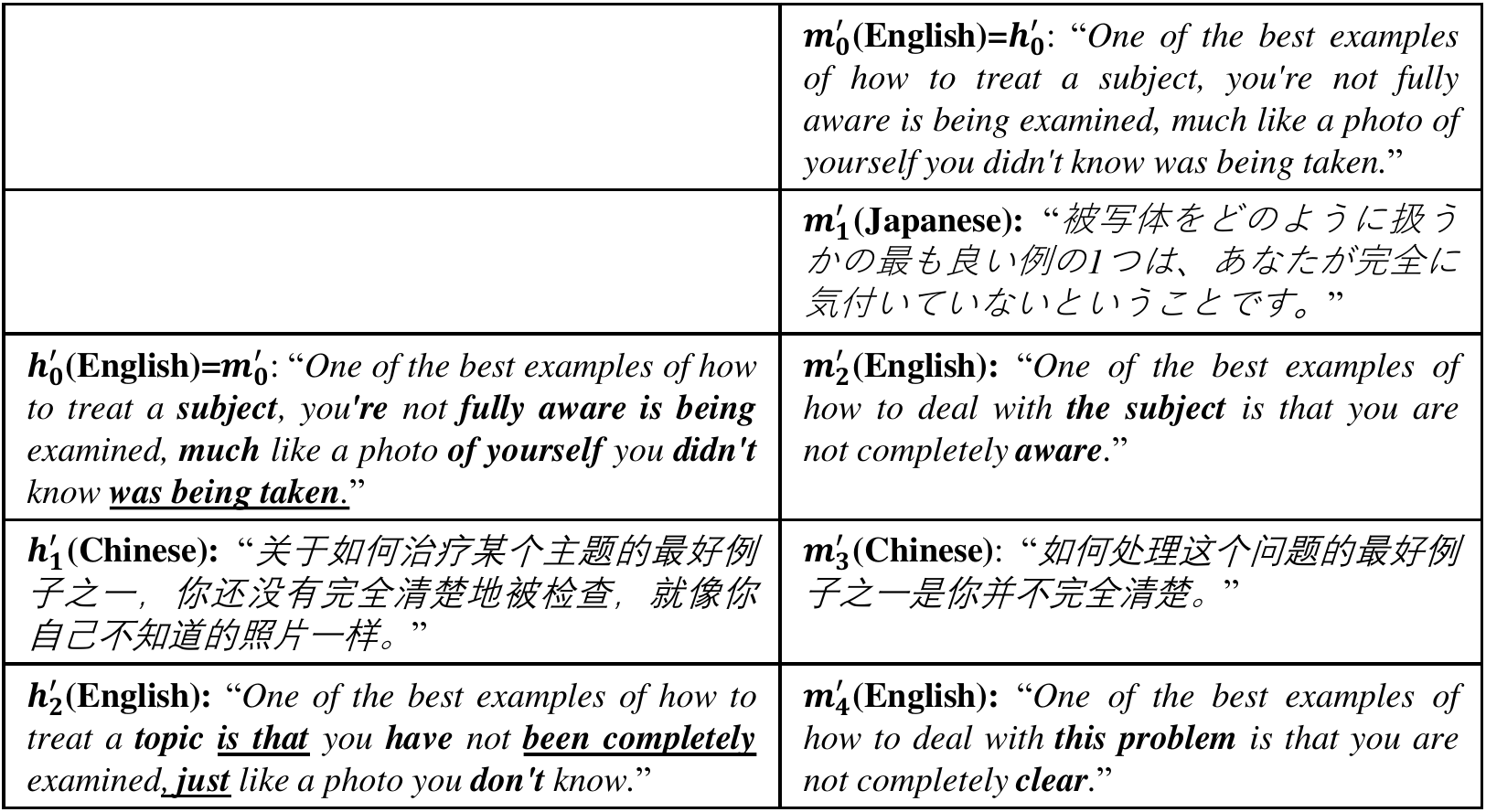}
\caption{Human vs machine text in back-translation detection.}
\label{Fig_3_Back_translation_detection}
\end{figure*}

Figure~\ref{Fig_3_Back_translation_detection} shows an example of back-translation detection.
In particular, we use Japanese for generating the machine-translated text $m'_2$ from the original text $m'_0$.
For distinguishing the two input texts $h'_0$ and $m'_2$, we create their back-translated texts $h'_2$ and $m'_4$ respectively with Chinese.
Like Figure~\ref{Fig_1_Example}, we highlight the variants between the input texts and their back-translations with bold for word usages and underline for structures.
Although using different languages in the generator and the detector, $h'_2$ still makes more variants than $m'_4$.
Again, the translation with four times in $m'_4$ causes fewer changes than that with two times in $h'_2$.


\subsection{Calculating Similarity (Step 2)}

\begin{figure}[t]
\centering
\includegraphics[]{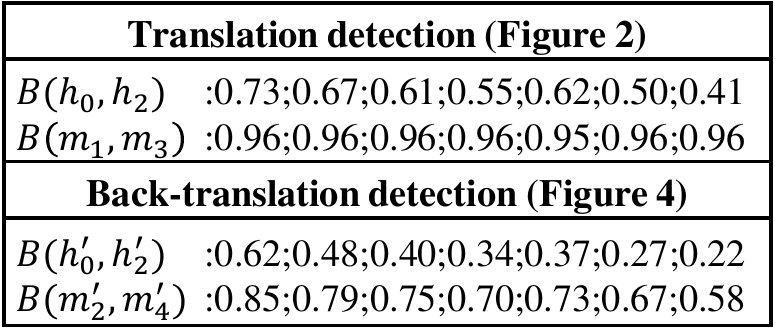}
\caption{BLEU scores $B$ of the human and machine texts with their back-translations.}
\label{Fig_4_BLEU_scores}
\end{figure}

This step aims to estimate the similarity between the input and its back-translation.
Due to the high relevance with machine translated-text measurement, we use BLEU scores~\cite{papineni2002bleu} for this step.
There are two groups of the BLEU including individual $N$-gram and cumulative $N$-gram scores.
While the individuals estimate phrases in the text independently, the later scores cumulate the measurements of the phrases with various lengths.
Because the individual uni-gram score equals to the cumulative uni-gram, we only use one of them.
The BLEU scores $B$ for both translation and back-translation detection are listed in Figure~\ref{Fig_4_BLEU_scores}.
The first four values indicate the individual $N$-gram with $N$ within 1 and 4; the remaining values are represented for the cumulative $N$-gram with $N$ from 2 to 4.

The results show that the BLEU scores between machine texts and their back-translations are all higher than those of machine texts in both translation and back-translation detection.
It demonstrates that the more times use a translator, the higher similarity is taken.
This significant information can be used to distinguish the human with machine text.

\subsection{Classifying the Input (Step 3)}

The seven BLEU scores extracted from the previous step are run with a classifier to determine whether the input text is translated by a machine or is written by a human.
We examine with four best classifiers, chosen from previous work, including linear classification~\cite{fan2008liblinear}, adaptive boosting, support vector machine (SVM) optimized by sequential minimal optimization, and SVM optimized by stochastic gradient descent.
All of the classifiers achieve nearly similar results, so we can use any of them for this step.

\section{Evaluation}
\label{section:evaluation}

\subsection{Translation Detection}

\subsubsection{Dataset}
\label{subsection:translation_dataset}
We randomly selected 2000 English-French sentence pairs from the European parallel corpus\footnote{\url{https://statmt.org/europarl/}}.
While the English was used as human-written texts, the French was translated to English by Google for producing machine texts.
The 4000 sentences are merged together; the integrated dataset contains 26.2 words per sentence on average.
The dataset then is split into two parts: 2800 sentences for a train set and the remaining for the test set.
To balance between human and machine texts in each set, we distribute both human and corresponding machine sentences into the same set.

\subsubsection{Comparison}

We evaluated the dataset on previous methods which detect machine-translated texts and machine-generated reviews.
The train set was learned with four machine learning classifiers, which were chosen as the best classifiers in the previous methods.
The topmost classifiers, which are mentioned in each method, are marked in underline as shown in Table~\ref{tab:translation_comparison}.
They include linear classification (LINEAR)~\cite{fan2008liblinear}, adaptive boosting (ADABOOST), support vector machine optimized by sequential minimal optimization SVM(SMO), and SVM optimized by stochastic gradient descent SVM(SGD).
Two standard metrics are used to evaluate the classifiers including accuracy (ACC) and $F$-score (F1) whose best performances are highlighted in bold.
The two last columns calculate the average and the two last rows show the results of our detectors.
The first one chooses Spanish for extracting back-translation information that is different from the language of the generator.
The second detector uses the same generator language, i.e. French.

\begin{table*}[t]
\begin{center}
\setlength\tabcolsep{2.0pt} 
\begin{tabular}{l c c c c c c c c | c c}

\hline 
\multirow{2}{*}{\textbf{Method}} &\multicolumn{2}{c}{\textbf{LINEAR}} & \multicolumn{2}{c}{\textbf{ADABOOST}}& \multicolumn{2}{c}{\textbf{SVM(SMO)}} & \multicolumn{2}{c|}{\textbf{SVM(SGD)}} &  \multicolumn{2}{c}{\textbf{AVERAGE}} \\ 
&\textbf{ACC} &\textbf{F1} &\textbf{ACC}&\textbf{F1}&\textbf{ACC}&\textbf{F1}&\textbf{ACC}&\textbf{F1}&\textbf{ACC}&\textbf{F1}\\ 
\hline

\makecell[l]{Word distribution\\\cite{nguyen2017identifying}} & \textbf{54.8\%}    &    \textbf{54.8\%}    &    53.8\%    &    53.0\%    &    52.8\%    &    52.6\%    &    \underline{52.0\%}    &    \underline{51.2\%}  &53.4\%	& 52.9\%\\
\hline 

\makecell[l]{Coherence\\\cite{nguyen2019detecting}} &55.0\%    &    53.3\%    &    53.8\%    &    41.6\%    &    \underline{\textbf{56.9\%}}    &    \underline{\textbf{56.8\%}}    &    50.2\%    &    50.0\%  & 54.0\%	& 50.4\%
\\
\hline 

\makecell[l]{Parsing tree\\\cite{li2015machine}} &\underline{\textbf{55.2\%}}    &    \underline{\textbf{55.0\%}}    &    53.7\%    &    52.4\%    &    54.8\%    &    54.2\%    &    54.2\%    &    54.2\% & 54.5\%	& 54.0\% \\
\hline

\makecell[l]{$N$-gram \& functional words\\\cite{aharoni2014automatic}}  &56.9\%    &    56.9\%    &    56.0\%    &    55.2\%    &    \underline{\textbf{58.2\%}}    &    \underline{\textbf{58.2\%}}    &    50.3\%    &    50.2\%  & 55.3\%	& 55.1\%\\
\hline 

\makecell[l]{$N$-gram \& readability\\\cite{juuti2018stay}} &\textbf{62.8\%}    &    \textbf{62.7\%}    &    \underline{52.1\%}    &    \underline{37.8\%}    &    61.5\%    &    61.5\%    &    55.1\%    &    55.1\% &57.9\%	& 54.3\% \\

\hline 

Our using Spanish  &	64.8\%    &    64.7\%    &    \textbf{65.6\%}    &    \textbf{65.5\%}    &    64.1\%    &    64.1\%    &    63.8\%    &    63.8\%  & 64.6\%	& 64.5\%\\

Our using French &	73.9\%    &    73.9\%    &    72.8\%    &    72.7\%    &    74.1\%    &    74.1\%    &    \textbf{75.0\%}    &    \textbf{75.0\%} &73.9\%	& 73.9\%
 \\

\hline 

\end{tabular}
\caption{Comparison with other methods on machine-translation detection.}
\label{tab:translation_comparison}
\end{center}
\end{table*}

Surprisingly, both accuracy and $F$-score of most previous methods are nearly the randomize approach which is around 50\%.
It demonstrates the balanced pairs in the dataset with mostly the same meanings and word usages between human and machine texts make the existing methods confused.
The best results are identical or comparative with the topmost performances mentioned in corresponding previous methods.
The huge difference comes from the \newcite{juuti2018stay}'s method because it is originally targeted on detecting another machine-generated text, namely fake reviews. 
Among them, the method based on word distribution~\cite{nguyen2017identifying} has the lowest results.
It indicates that the limited number of words within a sentence is insufficient to form a stable distribution.
The coherence-based method~\cite{nguyen2019detecting} appropriately targets on paragraph level but not in sentence level.
On the other hand, the method based on the parsing tree~\cite{li2015machine} can slightly improve the outcome but the structures of human and machine pairs seem to be similar in this balanced dataset.
The most reasonable methods~\cite{aharoni2014automatic,juuti2018stay} are to use $N$-gram model for measuring the text fluency.
However, the performances are unstable among classifiers especially in \newcite{juuti2018stay}'s work.
It indicates that the current neural translator has already improved, so the only use of internal text information is insufficient to recognize machine-translated texts.

Our method uses additional information from back-translation improving the overall performances.
Even using the different language with machine generator, the Spanish-based method achieves higher performance in all classifiers.
It demonstrates that our method can efficiently detect machine-translated text without information of the translation language.
Moreover, the use of the same language reaches the best performances in both accuracy and $F$-score.
Our method also gives a more balancing results not only between accuracy and $F$-score but also among classifiers.

\subsection{Back-translation Detection}

\subsubsection{Dataset}

We check the capability of the proposed method on another task, namely back-translation detection.
The back-translation can be easily used to generating paraphrasing texts for supporting malicious purposes such as fake reviews or political posts.
The generation needs only an original text and using back-translation with various languages for generating many paraphrasing versions.
For simulating this scenario, we randomly picked up 2000 sentiment sentences from Stanford Treebank corpus\footnote{\url{http://nlp.stanford.edu/~socherr/stanfordSentimentTreebank.zip}}.
Half of them is positive while the remaining one is negative.
We then generated the back-translated texts which are considered as machine sentences using French as the intermediate language.
The machine sentences were integrated with the original ones into 4000 sentence dataset that averagely has 17.3 words per sentence.
It is obviously smaller than the translation dataset above due to short sentences such as ``\textit{Imperfect?}'' and ``\textit{Cool}.''
The dataset was also split into train and test sets with balancing human and machine sentences in the same manner with the section~\ref{subsection:translation_dataset}.

\subsubsection{Comparison}

We conducted similar experiments with the previous methods on this back-translation dataset.
For evaluating our detectors, we also use two languages including Spanish and French.
The first language is different from the generator while the last is the same.
The results are shown in Table~\ref{tab:rich_resource_back_translation}.

\begin{table*}[t]
\begin{center}
\setlength\tabcolsep{2.0pt} 
\begin{tabular}{l c c c c c c c c | c  c}

\hline 
{\multirow{2}{*}{\textbf{Method}}} &\multicolumn{2}{c}{\textbf{LINEAR}} & \multicolumn{2}{c}{\textbf{ADABOOST}}& \multicolumn{2}{c}{\textbf{SVM(SMO)}} & \multicolumn{2}{c|}{\textbf{SVM(SGD)}} & \multicolumn{2}{c}{\textbf{AVERAGE}} \\ 
&\textbf{ACC} &\textbf{F1} &\textbf{ACC}&\textbf{F1} &\textbf{ACC}&\textbf{F1}&\textbf{ACC}&\textbf{F1}&\textbf{ACC}&\textbf{F1}\\ 
\hline

\makecell[l]{Word distribution\\\cite{nguyen2017identifying}}  & 53.8\%    &    53.5\%    &    \textbf{54.3\%}    &    \textbf{53.8\%}    &    53.6\%    &    52.8\%    &    \underline{51.4\%}    &    \underline{50.4\%}  &53.3\%&	52.6\%
\\

\hline

\makecell[l]{Coherence\\\cite{nguyen2019detecting}} & 51.2\%    &    37.1\%    &    58.3\%    &    57.4\%    &    \underline{\textbf{59.4\%}}    &    \underline{\textbf{59.4\%}}    &    49.3\%    &    49.3\%  & 54.5\% &	50.8\%
\\

\hline

\makecell[l]{Parsing tree\\\cite{li2015machine}}  & \underline{56.2\%}    &    \underline{56.2\%}    &    54.8\%    &    54.8\%    &    \textbf{56.6\%}    &    \textbf{56.6\%}    &    54.8\%    &    54.1\%  &55.6\%	& 55.4\%
\\

\hline

\makecell[l]{$N$-gram \& functional words\\\cite{aharoni2014automatic}}  & 57.7\%    &    57.7\%    &    55.9\%    &    52.5\%    &    \underline{\textbf{58.3\%}}    &    \underline{\textbf{58.3\%}}    &    45.4\%    &    45.3\%  &54.3\%&	53.5\%
\\

\hline

\makecell[l]{$N$-gram \& readability\\\cite{juuti2018stay}} & 61.9\%    &    57.5\%    &    \underline{\textbf{66.7\%}}    &    \underline{62.5\%}    &    63.8\%    &    \textbf{63.7\%}    &    56.3\%    &    56.3\%  &62.2\%	& 60.0\%
\\

\hline 

Our using Spanish  &	70.7\%    &    70.7\%    &    70.6\%    &    70.6\%    &    \textbf{71.0\%}    &    \textbf{71.0\%}    &    68.3\%    &    68.3\%  &70.1\%	&70.1\%
\\

Our using French &	83.0\%    &    83.0\%    &    83.1\%    &    83.0\%    &    83.1\%    &    83.0\%    &    \textbf{83.4\%}    &    \textbf{83.4\%}  &83.1\%	& 83.1\%
\\

\hline 

\end{tabular}
\caption{Back-translation detection with rich-resource language.}
\label{tab:rich_resource_back_translation}
\end{center}
\end{table*}

The performances on most previous methods are slightly increased.
The main reason is that the back-translation machine texts are generated after using the translator in two times.
Therefore, the quality is downgraded and this text is more easily distinguishable.
The most changing comes from \newcite{juuti2018stay}'s work in which the ADABOOST reaches the best accuracy contrasting with the result from the previous task in Table~\ref{tab:translation_comparison}.
Moreover, the best $F$-score places in another classifier, i.e., SVM(SMO).
The differences with other classifiers are also remarkable that exploits the inconsistent of this work on detecting various kinds of translated texts.
On the other hand, our methods achieve the highest performances in both Spanish and French.
The improvements are even larger compared with the previous task.
In this task, the back translations of machine texts are created after using the translator in four times while the previous task is only three, so we can exploit more differences comparing with the back-translations created by human texts in the same two times.
The proposed method again demonstrates the high consistent results among classifiers.
Furthermore, the accuracy and $F$-scores are almost identical.
It shows the persistent of our method on various tasks even without language information from the generator.

\subsubsection{Low-resource language}

\begin{table*}[t]
\begin{center}
\setlength\tabcolsep{2.0pt} 
\begin{tabular}{l c c c c c c c c | c c}

\hline 
{\multirow{2}{*}{\textbf{Method}}} &\multicolumn{2}{c}{\textbf{LINEAR}} & \multicolumn{2}{c}{\textbf{ADABOOST}}& \multicolumn{2}{c}{\textbf{SVM(SMO)}} & \multicolumn{2}{c|}{\textbf{SVM(SGD)}} & \multicolumn{2}{c}{\textbf{AVERAGE}} \\ 
&\textbf{ACC} &\textbf{F1} &\textbf{ACC}&\textbf{F1} &\textbf{ACC}&\textbf{F1}&\textbf{ACC}&\textbf{F1}&\textbf{ACC}&\textbf{F1}\\ 
\hline

\makecell[l]{Word distribution\\\cite{nguyen2017identifying}} & \textbf{52.4\%}    &    \textbf{52.3\%}    &    52.4\%    &    51.4\%    &    50.8\%    &    49.9\%    &    \underline{51.3\%}    &    \underline{50.7\%}  & 51.7\%	& 51.1\%
\\

\hline

\makecell[l]{Coherence\\\cite{nguyen2019detecting}} & 51.5\%    &    39.1\%    &    \textbf{58.8\%}    &    \textbf{58.8\%}    &    \underline{\textbf{58.8\%}}    &    \underline{\textbf{58.8\%}}    &    53.1\%    &    52.8\%  &55.6\%	&52.4\%
\\

\hline

\makecell[l]{Parsing tree\\\cite{li2015machine}} & \underline{57.8\%}    &    \underline{57.7\%}    &    57.6\%    &    57.6\%    &    \textbf{58.4\%}    &    \textbf{58.4\%}    &    54.6\%    &    54.5\%  & 57.1\%	& 57.1\%
\\

\hline

\makecell[l]{$N$-gram \& functional words\\\cite{aharoni2014automatic}}  & 55.6\%    &    55.6\%    &    53.5\%    &    43.0\%    &    \underline{\textbf{56.9\%}}    &    \underline{\textbf{56.9\%}}    &    47.4\%    &    47.4\%  &53.4\%	& 50.7\%
\\

\hline

\makecell[l]{$N$-gram \& readability\\\cite{juuti2018stay}} & \textbf{66.1\%}    &    \textbf{65.6\%}    &    \underline{55.8\%}    &    \underline{45.0\%}    &    65.6\%    &    65.5\%    &    58.0\%    &    58.0\%  & 61.4\%	& 58.5\%
\\

\hline 

Our using Chinese  &	\textbf{65.2\%}    &    \textbf{65.2\%}    &    64.6\%    &    64.2\%    &    63.9\%    &    63.9\%    &    63.3\%    &    63.3\%  &64.2\%	& 64.1\%
\\

Our using Japanese &	\textbf{80.6\%}    &    \textbf{80.6\%}    &    78.9\%    &    78.5\%    &    79.8\%    &    79.7\%    &    78.7\%    &    78.6\%  &79.5\%	& 79.3\%\\

\hline 

\end{tabular}
\caption{Back-translation detection with low-resource language.}
\label{tab:low_resource_back_translation}
\end{center}
\end{table*}

We examined similar experiments with low-resource languages.
With the same dataset of 2000 human sentiment sentences, we choose Japanese for generating machine back-translated texts.
For detectors, we use two languages as intermediate languages for generating back-translation.
They include Chinese, different from the generator, and the same language, i.e Japanese. 
The results of comparison with other methods are listed in Table~\ref{tab:low_resource_back_translation}.

In the previous methods, the results are quite similar to detecting back-translation with the rich-resource language.
The most difference again lays in \newcite{juuti2018stay}'s work.
The best performances come back to LINEAR while ADABOOST is dropped down to the lowest.
It indicates the inconsistent of this method not only on different tasks but also on the same task with different resource languages.
Comparing Figure~\ref{Fig_1_Example} and Figure~\ref{Fig_3_Back_translation_detection}, the translated texts from rich-resource have still higher qualities than those from low-resource.
Therefore, it affects the back-translation information which is used in our method.
Especially with Chinese, our detector is slightly lower than some classifiers of the previous work, but the stable outcome still demonstrates via average, which is better than the state-of-the-art methods.
Moreover, the experiment with Japanese outperforms in all classifiers with significant improvements of both accuracy as 14.5\% and $F$-score as 15.0\%.

\section{Conclusion}
\label{section:conclusion}

In this paper, we have exploited that when using machine translators many times, the translated text is converged.
Moreover, the variant between two consecutive usages gets to be smaller.
We then propose a method for estimating the variant by using BLEU scores and use them for detecting two types machine-generated text: machine translation and machine back-translation.
In machine translation detection, the evaluation of French sentences on best classifiers, which are mentioned on previous methods, shows that our method can detect translated text with 75.0\% of both accuracy and $F$-score.
It outperforms the previous methods with the best accuracy as 62.8\% and $F$-score as 62.7\%.
In back-translated text detection, the performance is even significantly improved from 66.7\% to 83.4\% of accuracy and from 63.7\% to 83.4\% of $F$-score.
The experiments on low-resource language, i.e., Japanese, achieve similar results.
Moreover, we conduct similar experiments with different languages between generators and detectors.
Although the performances are lower than the same language experiments, our detectors still are better than the existing work in all classifiers related to rich-resource languages and are higher on average performances of the classifiers with low-resource.
It demonstrates our detectors work well even without language information of the generators.

In future work, we will investigate the effect of our findings on various translators such as neural-network-based and phrase-based translators.
Moreover, we will verify the use of the same translator but trained on different corpora.
We also analyze other machine generators for detecting other malicious texts such as adversarial texts, artificial fake news, etc.
Beyond text, the applications using our hypothesis for detecting other machine-generated data (e.g., image, video, sound, and structured data) will be considered.

\bibliography{back_translation}
\bibliographystyle{acl_natbib}

\end{document}